\documentclass{article}
\usepackage[preprint]{spconf}
\usepackage{amsmath,graphicx,hyperref}

\usepackage[export]{adjustbox}
\usepackage{multicol}
\usepackage{multirow}
\usepackage{amssymb} 
\usepackage{arydshln} 
\usepackage{seqsplit} 

\usepackage{caption}
\usepackage{etoolbox} 

\toappear{
  \raisebox{-6mm}{
    \parbox[b]{\textwidth}{\footnotesize\raggedright
© 2026 IEEE. Personal use of this material is permitted. Permission from IEEE must be obtained for all other uses, in any current or future media, including reprinting/republishing this material for advertising or promotional purposes, creating new collective works, for resale or redistribution to servers or lists, or reuse of any copyrighted component of this work in other works.
}}}

\newcommand{\codeid}[1]{{\footnotesize\ttfamily\seqsplit{#1}}}

\title{Reading Between the Waves: \\Robust Topic Segmentation Using Inter-Sentence Audio Features}
\name{Steffen Freisinger, Philipp Seeberger, Tobias Bocklet, Korbinian Riedhammer}
\address{Technische Hochschule Nürnberg\\
    Keßlerplatz 12, 90489 Nürnberg, Germany\\
    firstname.lastname@th-nuernberg.de}
\begin{document}
\ninept
\maketitle
\begin{abstract}
Spoken content, such as online videos and podcasts, often spans multiple topics, which makes automatic topic segmentation essential for user navigation and downstream applications. 
However, current methods do not fully leverage acoustic features, leaving room for improvement. 
We propose a multi-modal approach that fine-tunes both a text encoder and a Siamese audio encoder, capturing acoustic cues around sentence boundaries. 
Experiments on a large-scale dataset of YouTube videos show substantial gains over text-only and multi-modal baselines. 
Our model also proves more resilient to ASR noise and outperforms a larger text-only baseline on three additional datasets in Portuguese, German, and English, underscoring the value of learned acoustic features for robust topic segmentation.
\end{abstract}
\section{Introduction}
\label{sec:intro}
Topic segmentation partitions a document into thematically coherent sub-segments, enabling smoother user navigation and laying the foundation for additional tasks, such as retrieval and summarization. 
While supervised approaches for topic segmentation have advanced considerably with the availability of large-scale annotated data~\cite{koshorek-etal-2018-text}, most methods focus on written text.
In contrast, the surge in multi-modal content - particularly audio and video recordings - presents new challenges: transcripts of spoken language are often informal, ungrammatical, and prone to automatic speech recognition (ASR) errors.
Yet, these recordings also offer the acoustic modality, which can provide useful cues for identifying topic boundaries.

Previous work has leveraged audio by extracting specific features (e.g., pitch~\cite{Soares-2019}) or by generating pretrained acoustic embeddings~\cite{berlage-2020,ghinassi-2023}. However, these methods do not incorporate end-to-end fine-tuning of the audio encoder.
Inspired by the success of fine-tuning pretrained transformers for text-only topic segmentation~\cite{lukasik-etal-2020-text,Ghosh-2022,retkowski-waibel-2024-text}, we extend this paradigm to the acoustic domain.
To align the acoustic features to the segmentation task, we jointly fine-tune text and audio encoders, enabling more robust topic segmentation in spoken documents.
While previous work encodes entire spoken sentences~\cite{berlage-2020,ghinassi-2023}, we hypothesize that acoustic cues for topic shifts (e.g., prosodic features or contextual sounds) are better found at the inter-sentence boundaries.
Our main contributions are as follows:
\begin{itemize}
    \vspace{-1mm}
    \itemsep0em 
    \setlength{\leftskip}{-1em}   
    \item We propose a multi-modal method that outperforms a larger text-only baseline, indicating that integrating audio is more effective than simply scaling the model.
    \item We show that end-to-end training and boundary-focused context improve the audio integration baseline by 5.37 \(F_1\).
    \item We evaluate ASR error and unseen-language effects, showing audio integration boosts robustness and generalization.
\end{itemize}

\section{Related Work}
\label{sec:related}
The task of topic segmentation is typically treated as a sequence tagging task, where every sentence in a document is labeled as segment boundary or not. 
\cite{koshorek-etal-2018-text} employ this technique and train a hierarchical model architecture on a large English dataset of Wikipedia articles, named \textsc{Wiki-727k}.
Their architecture stacks two LSTMs: one for encoding the sentences and a second one for predicting the sentence-level topic changes.
In contrast, \cite{lukasik-etal-2020-text} classify each inter-sentence boundary independently, feeding both the left and right context into a BERT-based model.
To efficiently encode long documents, \cite{abbasi-etal-2025-neural} introduce a sliding window approach for the document encoder.
\cite{Yu-2025} improve video topic segmentation by exploring cross-attention/MoE fusion architectures, using visual and text features.
\cite{Ghosh-2022} explore pretraining on \textsc{Wiki-727k} and find that it does not boost performance on a smaller dataset of chat transcripts.
However, \cite{retkowski-waibel-2024-text} show that pretraining on \textsc{Wiki-727k} can improve segmentation quality on their newly released \textsc{YTSeg} benchmark. 
Among unsupervised algorithms, TextTiling~\cite{hearst-1997-text} represents one of the earliest ones and detects topic boundaries by measuring lexical shifts across sentences.
Later variants employ pretrained text encoders \cite{Solbiati-2021} or hyperdimensional vectors~\cite{Park-2023}. 
Recent methods use LLMs for segmentation, by prompting the model to generate topic lists \cite{fan-etal-2024-uncovering}, or table-of-contents \cite{Freisinger-2025}.
However, the approaches mentioned above do not make full use of the audio modality.

When segmenting spoken documents (e.g., meeting or lectures recordings), the audio modality is often combined with textual transcripts.
For multi-participant dialogues, \cite{Zhang-2019} show that adding speaker information improves segmentation accuracy.
\cite{berlage-2020} concatenate acoustic and textual word embeddings for radio show segmentation, finding that embeddings pretrained on non-speech sound classification work best.
\cite{ghinassi-2023} extend this approach to magazine-style radio and news bulletins, comparing X-Vectors~\cite{snyder-2018} and L\textsuperscript{3}-Net~\cite{Cramer-2019}, with the latter yielding better performance.
Yet, these approaches only use frozen audio encoders without fine-tuning.

Apart from topic segmentation, self-supervised speech models have shown strong results in audio classification tasks.
Wav2vec 2.0 is pretrained by masking waveforms and predicting latent targets on 960 h of unlabeled audio~\cite{baevski-2020}.
Many studies fine-tune it for downstream tasks, e.g. vocal-intensity classification \cite{kodali-2024} or accent detection~\cite{ozturk-2024}.
HuBERT follows a similar pretrain and fine-tune recipe~\cite{hsu-2021} and has been adapted for tasks such as speech emotion recognition (SER)~\cite{gao-2024}.
UniSpeech-SAT is a speaker-aware extension of HuBERT that adds utterance-level contrastive objectives and utterance-mixing during self-supervised pretraining~\cite{Chen-2021}.
Prior work shows it performs strongly on speaker-oriented classification, keyword spotting, and SER, often on par with or better than Wav2Vec 2.0/HuBERT~\cite{Phukan-2023}.
These results motivate our experiments to align the audio encoders to the topic segmentation task.

\section{Methods}
\label{sec:methods}

\subsection{Siamese boundary audio encoder (inter-sentence context)}
\label{sec:audio-encoder}
Prior work typically encodes the \emph{sentence context}: the full audio of sentences \(n\) is passed once through a pretrained speech encoder and pooled to a single vector used by the topic segmenter~\cite{berlage-2020,ghinassi-2023}.
We aim to capture acoustic cues right at sentence boundaries. 
For each boundary between sentences \(n\!-\!1\) and \(n\), we extract two short windows of fixed duration \(\tau\): one at the end of \(n\!-\!1\) and one at the start of \(n\).
If the gap is shorter than \(\tau\), the windows meet at the midpoint (no overlap). 
Values for \(\tau\) are compared in \autoref{sec:ablations}.
This approach concentrates on acoustic cues present at the boundaries, such as prosodic changes, scene/speaker changes, and other sounds. 

\textbf{Siamese design (shared weights).}
Both boundary windows are encoded by the \emph{same} pretrained audio model \(f_\theta\) (Siamese). 
From each branch we take the final hidden states, mean-pool over time, and pass the pooled vector through a linear projector to \(d_{\text{proj}}=192\) dims. 
Concatenating left and right gives the acoustic \emph{boundary} feature
\(\mathbf{z}_n=\tanh([\mathbf{v}^L_n;\mathbf{v}^R_n])\in\mathbb{R}^{d_{\text{aud}}}\) with \(d_{\text{aud}}=2\,d_{\text{proj}}=384\). 
We chose \(d_{\text{aud}}\) to match the dimension of the sentence-level text features (described in next \autoref{sec:multiseg}).
Siamese networks are commonly used for tasks like spoof detection~\cite{Xie-2021} or for representation learning~\cite{reimers2019sentencebert}.
We use it for shared-weight acoustic feature extraction within our segmentation network.
The two-window setup cuts the encoder's attention cost versus a single \(2\tau\) window, and enables parallel processing.

\textbf{Fine-tuning.} 
We train the whole tagger model end-to-end; in our main model we fine-tune \(f_\theta\), and ablations compare fine-tuning vs. freezing (\autoref{sec:ablations}). 
For audio encoders, we use pretrained wav2vec 2.0~\cite{baevski-2020}, HuBERT~\cite{hsu-2021}, and UniSpeech-SAT~\cite{Chen-2021} models.
Whisper's encoder~\cite{Radford-2022} is also common in audio classification, but it expects 30 s inputs by design. 
With our 2 s boundary windows this causes heavy padding and large compute overhead, so it is impractical for our setting.

\subsection{MultiSeg}
\label{sec:multiseg}

\begin{figure}[t]
    \centering
    \includegraphics[width=0.97\linewidth]{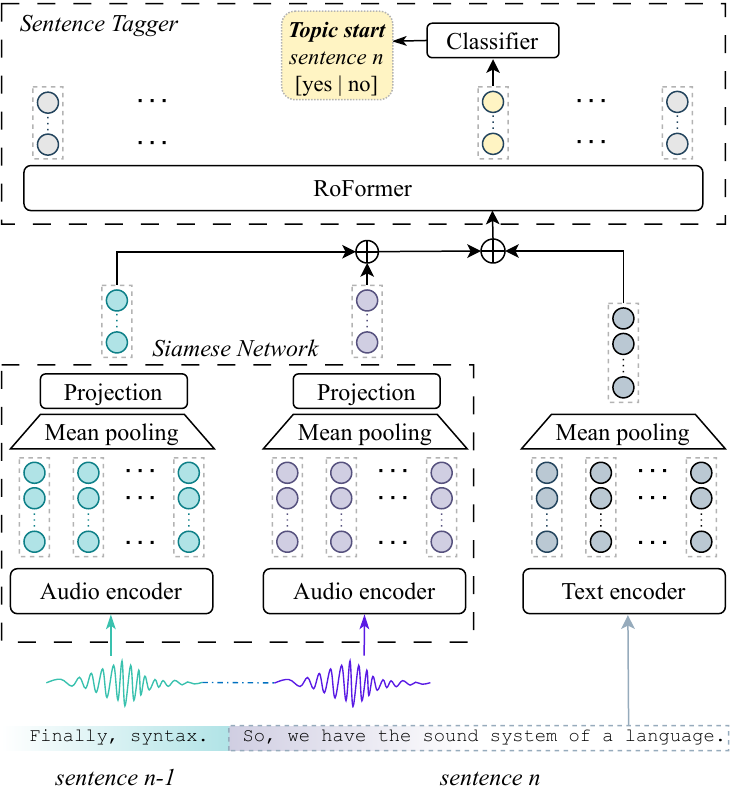}
    \vspace{-1mm}
    \caption{Our multi-modal topic change architecture.} 
    \label{fig:architecture}
\end{figure}

\textbf{Overview.} MultiSeg extends MiniSeg~\cite{retkowski-waibel-2024-text} by concatenating a sentence embedding with an acoustic \emph{boundary} feature (\autoref{sec:audio-encoder}).
\autoref{fig:architecture} visualizes the MultiSeg architecture.
We encode each sentence \(n\) with MiniLM~\cite{Wang-2020} into \(\mathbf{s}_n\in\mathbb{R}^{d_{\text{text}}}\) (\(d_{\text{text}}{=}384\)).
Then, \(\mathbf{s}_n\) is augmented with the boundary feature \(\mathbf{z}_n\in\mathbb{R}^{d_{\text{aud}}}\) taken from the boundary between sentences \(n\!-\!1\) and \(n\) (first sentence uses \(\mathbf{z}_1{=}\mathbf{0}\)); 
concatenation yields \(\mathbf{x}_n=[\mathbf{s}_n;\mathbf{z}_n]\in\mathbb{R}^{d_{\text{in}}}\) with \(d_{\text{in}}=d_{\text{text}}+d_{\text{aud}}=768\). 
For text-only MiniSeg, \(\mathbf{x}_n{=}\mathbf{s}_n\). 

\textbf{Tagger.} A RoFormer encoder~\cite{Su-2024} processes \((\mathbf{x}_1,\ldots,\mathbf{x}_N)\) and outputs contextual states \(\mathbf{u}_n\) (we use 12 layers, 8 heads, FFN 2048; hidden size \(=d_{\text{in}}\)). 
We map each contextual state \(\mathbf{u}_n\) to a single output logit \(y_n\) with a classifier module consisting of two hidden layers and ReLU activation. 
We compute probabilities \(p_n=\sigma(y_n)\) using Sigmoid and minimize binary cross-entropy \(\mathrm{BCE}(p_n,\, y_n^\ast)\) with targets \(y_n^\ast \in \{0,1\}\), where \(y_n^\ast=1\) means sentence \(n\) starts a new topic, and \(y_n^\ast=0\) means no topic change.

\textbf{Fusion and matching.} 
For every sentence \(n\), the matched acoustic feature \(\mathbf{z}_n\) comes from the \emph{inter-sentence} boundary \((n\!-\!1,n)\) (\autoref{sec:audio-encoder}).
This anchors audio cues exactly where segmentation decisions are made.

\section{Experiments}
\label{sec:experimental}

\begin{table*}[t]
  \caption{Topic segmentation results for \textsc{YTSeg} test set (\includegraphics[height=1em, valign=c]{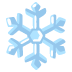} = frozen encoder, \includegraphics[height=1em, valign=c]{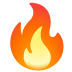} = fine-tuned encoder). Scores ± standard deviations are calculated by bootstrapping the test set 100 times, as done by \cite{lukasik-etal-2020-text}. (*metrics as reported by authors)}
  \label{tab:results}
  \centering

  \renewcommand{\arraystretch}{1.15}
  \small
  \begin{tabular}{lcc|ccccc}
    \hline
    \textbf{Method}  & \multicolumn{2}{c|}{\textbf{Features}} & $F_1\uparrow$ & $Prec\uparrow$ & $Rec\uparrow$ & $P_k\downarrow$ & $B\uparrow$ \\
      & Text & Audio &  &  &  &  & \\
    \hline
    ChatGPT \cite{fan-etal-2024-uncovering} & \includegraphics[height=1em, valign=c]{snowflake_emoji.png} & - &  39.16 ± 1.03 & 44.15 ± 1.53 & 35.20 ± 0.98  & 29.54 ± 0.46 & 33.79 ± 0.92 \\
    Cross-segment BERT \cite{lukasik-etal-2020-text} & \includegraphics[height=1em, valign=c]{fire_emoji.png} & - &  48.41 ± 0.94 & 46.91 ± 1.13  & \underline{50.02} ± 1.10  &  26.47 ± 0.48 &  40.72 ± 0.94 \\
    MiniSeg \cite{retkowski-waibel-2024-text}* & \includegraphics[height=1em, valign=c]{fire_emoji.png} & - &  43.37 ± 0.60 & 45.44 ± 0.83 & 41.48 ± 0.85  &  28.73 ± 0.39 &  35.74 ± 0.68 \\
    MiniSeg\textsuperscript{+} cf. \cite{retkowski-waibel-2024-text} & \includegraphics[height=1em, valign=c]{fire_emoji.png} & - &  \underline{48.83} ± 0.96 & \underline{51.87} ± 1.13 & 46.13 ± 1.09  & \underline{25.91} ± 0.46 &  \underline{41.17} ± 0.99 \\
    \hline
    MiniSeg +L\textsuperscript{3}-Net cf. \cite{ghinassi-2023} & \includegraphics[height=1em, valign=c]{fire_emoji.png} & \includegraphics[height=1em, valign=c]{snowflake_emoji.png}  & 47.61 ± 0.89 & 47.58 ± 0.84  & 47.65 ± 1.18 & 27.17 ± 0.48 &  37.75 ± 0.99 \\
    MultiSeg (ours) & \includegraphics[height=1em, valign=c]{fire_emoji.png} & \includegraphics[height=1em, valign=c]{fire_emoji.png} & \textbf{52.98} ± 0.93  & \textbf{52.77} ± 0.89  & \textbf{53.19} ± 1.18  &  \textbf{23.93} ± 0.50 &  \textbf{45.09} ± 1.02 \\
    \hline
  \end{tabular}
\end{table*}

\begin{table}[t]
  \caption{Model ablations (FT = Fine-tuned audio encoder).}
  \label{tab:ablations}
  \centering

  \setlength{\tabcolsep}{3pt}
  \renewcommand{\arraystretch}{1.06} 
  \small
  \begin{tabular}{lllcc|cc}
    \hline
    \textbf{Text Enc.} & \textbf{Ac. Ctx}   & \textbf{Audio Enc.} & \textbf{FT}  & \textbf{Dur} &  $F_1\uparrow$ & $B\uparrow$ \\ 
    \hline
    \multirow{9}{*}{MiniLM} & \multirow{7}{*}{Inter-sent.} & \multirow{5}{*}{wav2vec 2.0} & \checkmark & 1 s  & 52.63  & 44.36  \\ 
      &              &       & \checkmark & 2 s  & 52.98  & \textbf{45.09}  \\ 
      &              &       & \checkmark & 3 s  &  \textbf{53.29} &  44.52 \\ 
      &              &       & \checkmark & 4 s  & 52.87  &  45.03 \\ 
      \cline{4-7}
      &              &       & $\times$ & 2 s  & 51.19  &  42.33 \\ 
      \cline{3-7}
      &              & \multirow{2}{*}{HuBERT} & \checkmark    & 2 s  & 52.84  & 44.23  \\ 
      &              &      & $\times$ & 2 s  & 51.68  & 43.42  \\ 
      \cline{3-7}
      &              & \multirow{2}{*}{UniSpeech-SAT} & \checkmark    & 2 s  & 52.20  & 44.65  \\ 
      &              &      & $\times$ & 2 s  & 51.23  & 42.92  \\ 
      \cline{2-7}
      & \multirow{2}{*}{Sentence}
         & \multirow{2}{*}{wav2vec 2.0}
           & \checkmark & $\leq$8 s  &  51.02 &  43.24 \\ 
      &              &      
           & $\times$    & $\leq$8 s  &  47.57 & 40.14  \\ 
    \hline
    – & Inter-sent. & wav2vec 2.0   & \checkmark & 2 s  & 34.89  &  24.06 \\ 
    \hline
  \end{tabular}
\end{table}

\subsection{Experimental setup}
\textbf{Dataset.}
We conduct our experiments on the \textsc{YTSeg} dataset~\cite{retkowski-waibel-2024-text}, which contains 19{,}299 English YouTube videos from 393 channels. 
The videos span diverse topics, including science, lifestyle, politics, health, economy, and technology, and cover formats such as podcasts, lectures, and news. 
The authors provide fixed training, validation, and test splits, which we adopt without change. 
We use the validation set for early stopping and report results on the test set. 
As preprocessing step, we temporally align the transcripts and audio files using the Aeneas~\cite{aeneas} toolkit. 

\textbf{Training parameters.} For MiniSeg and our MultiSeg variants we follow the MiniSeg training recipe~\cite{retkowski-waibel-2024-text}. 
We optimize BCE on logits with AdamW (lr \(2.5\!\times\!10^{-5}\)), effective batch size of \(16\) videos (via gradient accumulation), and dropout \(0.1\). 
We further follow their gradient-sampling scheme to reduce memory usage and improve regularization. In the text-only setup, only half of the training samples backpropagate through the sentence encoder. 
In our multi-modal variant, for each sample, we choose whether gradients flow through the text encoder or the audio encoder (\(p=0.5\) each).
Training is run end-to-end on the BCE loss (see \autoref{sec:multiseg}); the RoFormer tagger and classifier head are always updated.
MiniSeg uses a positive-class weight of 2.0; for our multi-modal methods, we use a weight of 3.0, as we found it yields more balanced results here.
For training Cross-segment BERT, we follow the setup proposed by~\cite{lukasik-etal-2020-text}.

\textbf{Evaluation.}
To evaluate performance, we measure precision, recall, and \(F_1\) at the sentence boundary level.
Additionally, we report two segmentation-specific metrics, $P_k$~\cite{Beeferman1999StatisticalMF} and Boundary Similarity (\(B\))~\cite{fournier-2013-evaluating}.
These segmentation metrics are computed for each video and then aggregated using a weighted average based on the number of segments in each test sample.
We omit the label of the first sentence in all evaluations, as it is always set to zero (no topic change).

\subsection{Baselines}
We include three text-based baseline methods. 
The first, \textbf{Cross-segment BERT}~\cite{lukasik-etal-2020-text}, assigns a binary label (\textit{topic change} or \textit{no change}) at each inter-sentence boundary. 
For classification, it takes 128 tokens on both sides of the boundary, joins them with a $[SEP]$ token, and employs a classifier on top of the $[CLS]$ embedding.

The second approach, \textbf{MiniSeg}~\cite{retkowski-waibel-2024-text}, treats segmentation as a sequence-tagging task. 
It encodes sentences using a MiniLM encoder~\cite{Wang-2020} and tags them using a RoFormer~\cite{Su-2024} as \emph{topic change} or \emph{no change}. 
It has shown strong performance on the \textsc{Wiki-727k} dataset and sets the state-of-the-art for the \textsc{YTSeg} dataset~\cite{retkowski-waibel-2024-text}.

Our third baseline uses \textbf{ChatGPT}~\cite{fan-etal-2024-uncovering} in a zero-shot setting. 
They prompt the model to generate lists of sentence IDs, where each list represents one topic, following the approach in \cite{fan-etal-2024-uncovering}. 
Specifically, we run our experiments with the \textit{GPT-4o-mini} variant.

As a multi-modal baseline, we adopt the method of \cite{ghinassi-2023}, which is a closely related prior approach to multi-modal (text+audio) topic segmentation. 
They encode the audio of entire sentences using L\textsuperscript{3}-Net~\cite{Cramer-2019} and aggregate the frame-level features by concatenating the mean and standard deviation vectors, resulting in a 1024-d embedding per sentence.
We adopt this audio encoding in the MultiSeg architecture by fusing text and audio embeddings, as described in \autoref{sec:multiseg}, and refer to the model as \textbf{MiniSeg +L\textsuperscript{3}-Net}.
This ensures a fair comparison between the two ways of incorporating audio, making sure that the text encoder path stays equal.

\subsection{Main results}
\label{sec:general-results}

\textbf{Models.}
\autoref{tab:results} compares our method with various baselines.
We configure MultiSeg with a 2 s inter-sentence context and a wav2vec 2.0 encoder incl. fine-tuning, since this setup led to the best results in \autoref{sec:ablations}.
In addition to the standard MiniSeg model~\cite{retkowski-waibel-2024-text}, we include a scaled-up variant that replaces the MiniLM-based sentence encoder (\codeid{sentence-transformers/all-MiniLM-L6-v2}) with a pretrained RoBERTa model~\cite{Liu-2019} (\codeid{all-roberta-large-v1}), increasing parameters from 49 M to 457 M (M=million). 
We refer to this model as \textbf{MiniSeg\textsuperscript{+}}. 
This setup ensures a fair comparison, since our multi-modal architectures have more parameters due to the audio encoder.
Cross-segment BERT, MiniSeg +L\textsuperscript{3}-Net, and MultiSeg have 108 M, 189 M, and 183 M parameters, respectively.

\textbf{Results.}
Our model outperforms all text-only baselines, including the scaled-up MiniSeg\textsuperscript{+} by 4.15 on \(F_1\) and 3.92 on \(B\), while using 59.9\% fewer parameters.
This result underlines the value of incorporating audio features instead of merely enlarging the text encoder.
When compared to the L\textsuperscript{3}-Net-based audio encoding method~\cite{ghinassi-2023}, our model improves \(F_1\) by 5.37 and \(B\) by 7.34.

\textbf{Acoustic cues (case analysis).}
We inspected 15 videos with the largest gains of MultiSeg over MiniSeg\textsuperscript{+}. 
At topic boundaries we often observed localized cues such as brief pauses, pitch dips, emphatic restarts, speaker/scene switches, and short transition music/sound effects. 
Performance decreased when cues were not boundary-specific (e.g., recurring swoosh/ambient sounds throughout) or when boundary windows contained little signal (continuous speech), suggesting audio helps most when boundary cues are distinctive.

\subsection{Method ablations}
\label{sec:ablations}
\autoref{tab:ablations} reports the effect of audio-context design, encoder choice, and fine-tuning for our multi-modal setup.
We compare our inter-sentence acoustic context approach (presented in \autoref{sec:audio-encoder}) with a full sentence acoustic context. 
Our approach (2 s variant) outperforms the sentence-based context by 1.96 \(F_{1}\) and 1.85 \(B\).
Fine-tuning the acoustic encoder is crucial: keeping wav2vec 2.0~\cite{baevski-2020} (\codeid{facebook/wav2vec2-base}) frozen lowers performance by 1.79 \(F_{1}\) and 2.76 \(B\).
We further add an audio-only variant, dropping the text branch entirely.
The results show that using audio alone is substantially worse than all multi-modal variants, indicating that acoustic cues complement textual features rather than replace them.

Changing the backbone from wav2vec 2.0 to HuBERT~\cite{hsu-2021} (\codeid{facebook/hubert-base-ls960}) or UniSpeech-SAT~\cite{Phukan-2023} (\codeid{microsoft/unispeech-sat-base-plus}) produces a
comparable but slightly weaker system. 
Varying the boundary-window length shows diminishing returns beyond three seconds: \(F_{1}\) peaks for 3 s, while \(B\) is highest with 2 s.
We choose the 2 s variant, as it is more efficient and shows similar performance.
We further explored cross-modal fusion (e.g., cross-attention), but observed no consistent gains over simple concatenation.
Overall, the best trade-off is achieved with a 2 s inter-sentence context and a fine-tuned wav2vec 2.0 encoder. 
We use this configuration for all other MultiSeg experiments.

\begin{table}[t]
  \caption{Results on \textsc{YTSeg} test set, transcripts generated using different ASR models ($\Delta$ shows difference to \emph{Oracle} scores).}
  \label{tab:asr}
  \centering

  \setlength{\tabcolsep}{3pt}
  \renewcommand{\arraystretch}{1.1} 
  \small
  \begin{tabular}{l|c|cc|cc}
    \hline
     \textbf{ASR Model} & \textbf{WER} (\%) & \multicolumn{2}{c}{\textbf{MiniSeg\textsuperscript{+}}} & \multicolumn{2}{c}{\textbf{MultiSeg}}  \\
     
      & & \((\Delta) F_1 \) & \((\Delta)B  \) & \((\Delta)F_1  \) & \((\Delta)B \)  \\

    \hline
    Oracle &  - &  48.83 & 41.17 & 52.98 & 45.09 \\
    \hline
    Whisper large-v3 &  19.64 & -4.08 & -3.69 & -2.35 & -3.03 \\
    Whisper medium & 20.23 & -4.08 & -4.43 & -2.38 & -3.04 \\
    Whisper small & 20.78 & -4.65 & -4.95 & -2.54 & -3.24 \\
    Whisper base & 22.66 & -5.46 & -5.72 & -2.56 & -3.34 \\
    Whisper tiny & 24.88 & -5.78 & -6.27 & -2.57 & -3.72 \\
    Vosk-small & 38.13 & -12.75 & -13.23 & -5.83 & -7.08 \\
    \hline
  \end{tabular}
\end{table}

\subsection{Robustness to ASR errors}
\label{sec:wer}

The preceding experiments use the original transcripts released with \textsc{YTSeg}.
In real-world conditions, transcripts are usually produced by ASR systems, whose word error rate (WER) can vary widely.
We therefore assess how segmentation quality degrades when the input transcript is noisy.

\textbf{ASR systems.}
We transcribe every test video with six off-the-shelf ASR models.
Five are Whisper variants, which share the same Transformer architecture and differ in model size (39 M to 1550 M parameters)~\cite{Radford-2022}. 
For a lightweight baseline we add a 90 M-parameter Kaldi TDNN-F model that runs in real time on CPU~\cite{vosk-toolkit} (\codeid{vosk-model-small-en-us-0.15}).
WER is measured after punctuation stripping and lower-casing.

\textbf{Results.}
\autoref{tab:asr} reports the absolute scores on the original transcripts (\emph{Oracle}) and the score drops (\(\Delta\)) on ASR transcripts. 
For comparison, we use the text-only MiniSeg\textsuperscript{+} and the multi-modal MultiSeg, as presented in \autoref{sec:general-results}.
As WER grows from 19.6\% (Whisper large-v3) to 38.1\% (Vosk), both models degrade, but the multi-modal system loses less:
At 38\% WER the drop is \(5.8\,F_{1}\) versus \(12.8\,F_{1}\) for the text-only baseline.
The same pattern holds for \(B\).
This indicates that the additional audio modality might mitigate the impact of erroneous text input.

\subsection{Cross-dataset and cross-lingual evaluation}

We further evaluate our MultiSeg method on three additional datasets. 
Two of these datasets are in a language not seen during topic segmentation training.
We hypothesize that certain acoustic patterns (e.g., prosodic traits or nonverbal sounds) are less language-dependent than lexical cues, potentially improving robustness across different languages.

\textbf{Datasets.} While the \textsc{YTSeg} train set is used for training, we  evaluate on the following three datasets, each providing topic segment annotations:
(i) \textsc{AVLectures}~\cite{darshan-2023}, a dataset containing 350 annotated lecture videos;
(ii) \textsc{Videoaula}~\cite{Soares-2018}, a corpus of 34 Portuguese lecture videos covering computer science;
(iii) \textsc{LectureDE}~\cite{Freisinger-2023}, consisting of 95 German lecture videos covering engineering, social sciences, and didactics~\cite{Ranzenberger-2024}.
For all three test datasets, we follow the same preprocessing procedure: 
(i) transcribing using Whisper large-v2~\cite{Radford-2022};
(ii) synchronizing audio and transcript using the Montreal Forced Aligner~\cite{McAuliffe-2017};
(iii) segmenting transcript into sentences using SpaCy~\cite{honnibal-2020}.
This pipeline approximates real-world conditions, in which speech is often transcribed and split into sentences automatically.

\textbf{Models.} Our original models presented in \autoref{sec:general-results} are based on text encoders pretrained on English data.
While these English-only encoders suffice for the \textsc{AVLectures} corpus, they are unlikely to capture the Portuguese and German transcripts.
Accordingly, we retrain both MiniSeg and MultiSeg with multilingual sentence encoders on the English \textsc{YTSeg} data:
the text-only model adopts a pretrained MPNet-base encoder \cite{Song-2020} (\codeid{paraphrase-multilingual-mpnet-base-v2}), and the multi-modal one uses a multilingual MiniLM encoder \cite{Wang-2020} (\codeid{paraphrase-multilingual-MiniLM-L12-v2}). 
The models are named \textbf{Ml. MiniSeg\textsuperscript{+}} with 345 M parameters, and \textbf{Ml. MultiSeg} with 279 M parameters.

\textbf{Results.}
\autoref{tab:cross-results} shows that the multi-modal model outperforms
its larger text-only counterpart on each dataset.
On \textsc{AVLectures} (English) the \(F_{1}\) gap is modest
(\(+1.82\) points), but it widens on the non-English corpora:
\(+20.2\) \(F_{1}\) on Portuguese \textsc{Videoaula} and \(+6.93\) on
German \textsc{LectureDE}.
With regard to \(B\), the metrics behave similar.
These larger gains suggest that the acoustic modality provides language-independent signals that help the model locate topic boundaries when lexical cues are less reliable.
Overall, the experiment supports our hypothesis that multi-modal segmentation is more robust in cross-dataset and cross-lingual settings.

\begin{table}[t]
  \caption{Results on  cross-lingual datasets. Using multilingual (Ml.) model variants for the Portuguese and German datasets. Scores ± stddev calculated by bootstrapping the test set 100 times~\cite{lukasik-etal-2020-text}.}
  \label{tab:cross-results}
  \centering

  \setlength{\tabcolsep}{3pt}
  \renewcommand{\arraystretch}{1.2} 
  \small
  \begin{tabular}{lll|cc}
    \hline
    \textbf{Test set} & \textbf{Lang} & \textbf{Model} & $F_1\uparrow$ & $B\uparrow$ \\
    \hline
    \multirow{2}{*}{\textsc{AVLectures}} & \multirow{2}{*}{EN} & MiniSeg\textsuperscript{+} & 24.64 ± 1.67 &  15.97 ± 1.30 \\
      &    &  MultiSeg  & 26.46 ± 1.79 &  17.02 ± 1.40 \\
    \hline
    \multirow{2}{*}{\textsc{Videoaula}} & \multirow{2}{*}{PT} & Ml. MiniSeg\textsuperscript{+} & 30.39 ± 2.68 &  18.85 ± 2.01 \\
           &    &  Ml. MultiSeg  & 50.59 ± 3.14  &  33.58 ± 2.97 \\
    \hline
    \multirow{2}{*}{\textsc{LectureDE}} & \multirow{2}{*}{DE} & Ml. MiniSeg\textsuperscript{+} & 38.24 ± 3.15 &  25.72 ± 2.97 \\
       &    &  Ml. MultiSeg  & 45.17 ± 3.03 & 29.78 ± 3.22  \\
    \hline
  \end{tabular}
\end{table}

\section{Conclusion}
\label{sec:conclusion}
We introduced a multi-modal topic segmentation model that jointly fine-tunes text and audio encoders, focusing on inter-sentence boundary cues.
Experiments on the large-scale \textsc{YTSeg} dataset show notable gains over text-only and multi-modal baselines, highlighting the value of learning aligned acoustic features.
When using ASR-generated transcripts of different quality, we find that our multi-modal method is less prone to transcription errors.
Moreover, our cross-lingual evaluation on Portuguese and German datasets suggests that audio-enhanced models are more robust when applied to other languages.
We hope this research encourages broader exploration of topic segmentation in spoken content. 
Our model checkpoint and evaluation scripts can be found at: \url{https://github.com/steffrs/multimodal-topic-segmentation}.




\makeatletter
\patchcmd{\thebibliography}
  {\settowidth}
  {\setlength{\itemsep}{1.4pt}
   \setlength{\parsep}{0pt}
   \settowidth}{}{}
\makeatother

\begingroup
\fontsize{9}{10}\selectfont

\bibliographystyle{IEEEbib}
\bibliography{strings,refs}

\endgroup

\end{document}